# Joint auto-encoders: a flexible multi-task learning framework

**Baruch Epstein**  **Ron Meir**  **Tomer Michaeli**

The Viterbi Faculty of Electrical Engineering
Technion - Israel Institute of Technology, Haifa, Israel
baruch.epstein@gmail.com, rmeir@ee.technion.ac.il, tomer.m@ee.technion.ac.il

## Abstract

The incorporation of prior knowledge into learning is essential in achieving good performance based on small noisy samples. Such knowledge is often incorporated through the availability of related data arising from domains and tasks similar to the one of current interest. Ideally one would like to allow both the data for the current task and for previous related tasks to self-organize the learning system in such a way that commonalities and differences between the tasks are learned in a data-driven fashion. We develop a framework for learning multiple tasks simultaneously, based on sharing features that are common to all tasks, achieved through the use of a modular deep feedforward neural network consisting of shared branches, dealing with the common features of all tasks, and private branches, learning the specific unique aspects of each task. Once an appropriate weight sharing architecture has been established, learning takes place through standard algorithms for feedforward networks, e.g., stochastic gradient descent and its variations. The method deals with domain adaptation and multi-task learning in a unified fashion, and can easily deal with data arising from different types of sources. Numerical experiments demonstrate the effectiveness of learning in domain adaptation and transfer learning setups, and provide evidence for the flexible and task-oriented representations arising in the network.

## 1 Introduction

A major goal of inductive learning is the selection of a rule that generalizes well based on a finite set of examples. It has been known since early times [22], and quantified rigorously in precise terms (e.g., chapter 7 in [9]), that inductive learning is impossible unless some regularity assumptions are made about the world. Such assumptions, by their nature, go beyond the data, and are based on prior knowledge achieved through previous interactions with 'similar' problems. Following its early origins ([4, 13, 33]), the incorporation of prior knowledge into learning has become a major effort recently, and is gaining increasing success by relying on the rich representational flexibility available through current deep learning schemes [6]. Various aspects of prior knowledge are captured in different settings, such as learning-to-learn, lifelong learning, domain adaptation, transfer learning, multi-task learning, etc. (e.g., [16]). In this work, we consider the setup of multi-task learning, first formalized in [4], where a set of tasks is available for learning, and the objective of the learner is to extract knowledge from a subset of tasks in order to facilitate learning of other, related, tasks. Within the framework of representation learning [6], the core idea is that of shared representations, allowing a given task to benefit from what has been learned from other tasks, since the shared aspects of the representation are based on more information [39].

We consider both unsupervised and semi-supervised learning setups. In the former setting we have several related datasets, arising from possibly different domains, and aim to compress each dataset based on features that are shared between the datasets, and on features that are unique to each problem. Neither the shared nor the individual features are given apriori, but are learned using a deep neural network architecture within an auto-encoding scheme. While such a joint representation could, in principle, serve as a basis for supervised learning, it has become increasingly evident that representations should contain some information about the output (label) identity in order to perform well, and that using pre-training based on unlabeled data is not always advantageous (e.g., chap. 15 in [16]). However, since unlabeled data is far more abundant than labeled data, much useful information can be gained from it. We therefore propose a joint encoding-classification scheme where both labeled and unlabeled data are used for the multiple tasks, so that internal representations found reflect both types of data, but are learned simultaneously.

**The main contributions of this work are:** *(i)* A generic and flexible modular setup for combining unsupervised, supervised and transfer learning. *(ii)* Efficient transfer learning using mostly unsupervised data (i.e., very few labeled examples are required for successful transfer learning). *(iii)* End-to-end learning.

## 2   Related work

Previous related work can be broadly separated into two classes of models: *(i)* Generative models attempting to learn the input representations. *(ii)* Non-generative methods that construct separate or shared representations in a bottom-up fashion driven by the inputs.

We first discuss several works within the non-generative setting. The Deep Domain Confusion (DDC) algorithm in [35] studies the problems of unsupervised domain adaptation based on sets of unlabeled samples from the source and target domains, and supervised domain adaptation where a (usually small) subset of the target domain is labeled . By incorporating an adaptation layer and a domain confusion loss they learn a representation that optimizes both classification accuracy and domain invariance, where the latter is achieved by minimizing an appropriate discrepancy measure. By maintaining a small distance between the source and target representations, the classifier based on the small number of target labels makes good use of the relevant prior knowledge. The Deep Adaptation Network (DAN) in [27] is based on the observation that the feature transferability drops in higher layers of a deep network, and hence they propose to enhance the transferability in task-specific layers by reducing the domain discrepancy. This is achieved through embedding the task-specific layers into a reproducing kernel Hilbert space and matching the mean embeddings of different domains. The algorithm suggested in [11] is based on a standard feedforward deep network consisting of a feature extractor and a label predictor, augmented with a domain classifier that is connected to the feature extractor, and acts to modify the gradient during backpropagation. This adaptation promotes the similarity between the feature distributions in a domain adaptation task. The Deep Reconstruction Classification Network (DRCN) in [14] tackles the unsupervised domain adaptation task by jointly learning a shared encoding representation of the source and target domains based on minimizing a loss function that balances between the classification loss of the (labeled) source data and the reconstruction cost of the target data. The shared encoding parameters allow the target representation to benefit from the ample supervised data available to the source. In addition to these mostly algorithmic approaches, an increasing number of theoretical papers have attempted to provide a deeper understanding of the benefits available within this setting [5, 29].

Next, we mention some recent work within the generative approach. Since our focus in this work is on a non-generative approach, we limit our summary of this topic. Early work [39] offered a probabilistic generative framework for multi-task learning where common task parameters share a common structure through latent variables, while task-specific parameters are unique to each task. The authors make statistical assumptions about the latent variables allowing them to construct flexible families of models, which can be learned through an EM-like algorithm. More recent work has suggested several extensions of the increasingly popular Generative Adversarial Networks (GAN) framework [17]. The Coupled Generative Adversarial Network (CoGAN) framework in [26] aims to generate pairs of corresponding representations from inputs arising from different domains. They propose a way to learn joint distributions over two domains based only on samples from the marginals. While this is not possible in a fully general setting, it is possible in specific cases by constructing two GANs sharing high-level features for the two domains, constructed by sharing weights both within the generator and within the discriminator networks. The Adversarial Discriminative Domain Adaptation (ADDA) approach [34] subsumes some previous results within the GAN framework of domain adaptation. The approach learns a discriminative representation using the data in the labeled source domain, and then learns to adapt the model for use in the (unlabeled) target domain through a domain adversarial loss function. The idea is implemented through a minimax formulation similar to the original GAN setup. Other relevant work in this direction includes [10, 12].

Finally we briefly mention that a great deal of work has been devoted to multi-modal learning where the inputs arise from different modalities. Exploiting data from multiple sources (or views) to extract meaningful features, is often done by seeking representations that are sensitive only to the common variability in the views and are indifferent to view-specific variations. Many methods in this category attempt to maximize the correlation between the learned representations, as in the linear canonical correlation analysis (CCA) technique [20] and its various nonlinear extensions [2, 31, 38, 30, 3, 18]. Other methods use losses based on both correlation and reconstruction error (in an auto-encoding like scheme) [37], or employ diffusion processes to reveal the common underlying manifold [25]. However, all multi-view representation learning algorithms rely on *paired examples* from the two views. This setting is thus very different from transfer learning, multi-task learning, or domain adaptation, where one has access only to *unpaired samples* from each of the domains.

Our work falls into the class of non-generative approaches. While GANs and their extensions provide a powerful approach to multi-task learning and domain adaptation, they are often hard to train and fine tune (see discussion in [15]). Our approach offers a complementary non-generative perspective, as reviewed above, and operates in an end-to-end fashion allowing the parallel training of multiple tasks, incorporating both unsupervised, supervised and transfer settings within a single architecture. This simplicity allows the utilization of standard optimization techniques for regular deep feedforward networks, so that any advances in that domain translate directly into improvements in our results. The approach does not require paired inputs and can operate with inputs arising from entirely different domains, such as speech and audio (although this has not been demonstrated empirically here). Our work is closest to [14], but offers a more flexible approach in that it starts from multiple tasks/domains for which simple autoencoders can be constructed and combines them in an unsupervised or supervised setting by sharing weights while maintaining a common shared representation as well as private representations for each task. The modular structure allows sharing weights at any level of the hierarchy, and enables the system to learn effectively by simple SGD algorithms.

## 3  Joint autoencoders

In this section, we introduce *joint autoencoders*, a general method for multi-task learning by unsupervised extraction of features shared by the tasks as well as features specific to each task. We begin by presenting a simple case, point out the various possible generalizations, and finally describe two transfer learning procedures utilizing joint autoencoders.

### 3.1  Joint autoencoders for reconstruction

A popular approach to extracting low-level features from high-dimensional data is to use autoencoders (see Fig 1(a)). In these architectures, the bottleneck features constitute an informative compact representation of the data, in the sense that they allow to reconstruct the data with minimal error. Now, suppose we have two types of data, $X^1$ and $X^2$, from domains $\mathcal{X}^1$ and $\mathcal{X}^2$, respectively. We would like to jointly learn how to auto-encode those two sources, in a way which exploits the similarities between them. To do this, we make the following two observations:

*(i)* Certain aspects of the two auto-encoding tasks we are facing, may be similar, but other aspects may be completely different (e.g., when the two domains contain color images and grayscale images, respectively).

*(ii)* The similarity between the two tasks can be rather "deep". For example, cartoon images and natural images may benefit from different low-level features, but may certainly share high-level structures.

To accommodate these two observations, we propose to split each of the autoencoders into a *private* branch and a *common* branch, as illustrated in Fig. 1(b) . In the common branches, some of the weights are shared between the autoencoders, while the private branches (top and bottom branches in Fig 1(b)) do not share weights. The shared weights can occur anywhere along the hierarchy, and do not need to be limited to the shallow layers. We call this architecture a *joint autoencoder* (JAE). The key idea is that the weight sharing forces the common branches to learn to represent the common features of the two sources. Consequently, the private branches are implicitly forced to capture only the features that are not common to the other task. Figure 1 illustrates the construction of a JAE out of a pair of autoencoders for the MNIST dataset [24], inspired by [19]. Here we have split the digits into two subsets, $X^1 = \{0, 1, 2, 3, 4\}$ and $X^2 = \{5, 6, 7, 8, 9\}$. The resulting network can be trained with standard backpropagation on both reconstruction losses simultaneously.

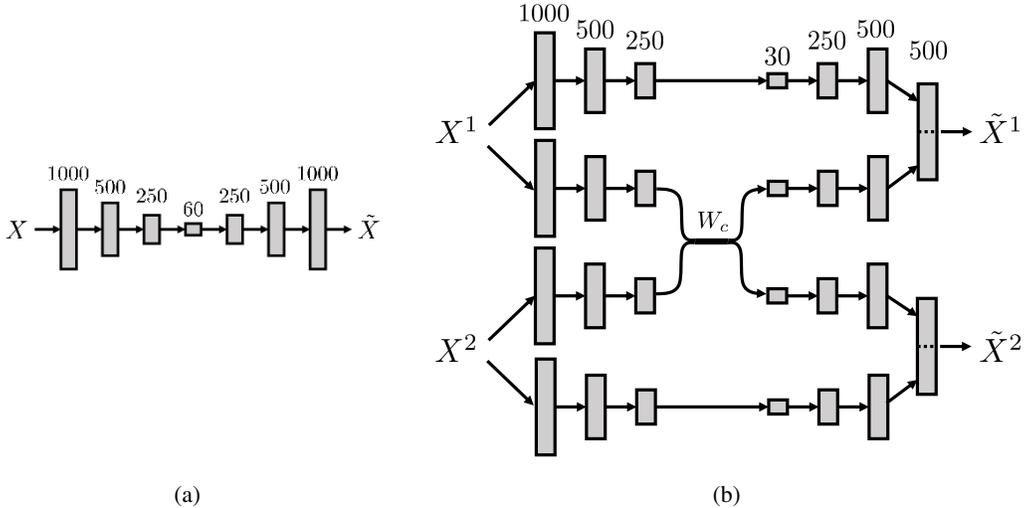

(a) (b)

Figure 1: An example of an MNIST autoencoder (a) and the joint autoencoder constructed out of it (b), where $X^1 = \{0, 1, 2, 3, 4\}$ and $X^2 = \{5, 6, 7, 8, 9\}$. Each layer is a fully connected one, of the specified size, with ReLU activations. The weights shared by the two parts are denoted by $W_c$. The pairs of the top fully connected layers of dimension 500 are concatenated to create a layer of dimension 1000 which is then used directly to reconstruct the input of size 784.

As mentioned before, in this simple example, both inputs are MNIST digits, all branches have the same architecture, and the bottlenecks are single layers of the same dimension. However, this need not be the case. The inputs can be entirely different (e.g., image and text), all branches may have different architectures, the bottleneck sizes can vary, and more than a single layer can be shared. Furthermore, the shared layers need not be the bottlenecks, in general. Finally, the generalization to more than two tasks is straightforward - we simply replace the autoencoder of each task by a pair of autoencoders with smaller bottlenecks and share some of the layers of the common-feature autoencoders. Weight sharing can take place between subsets of tasks, and can occur at different levels for the different tasks.

### 3.2 Joint autoencoders for multi-task semi-supervised and transfer learning

Consider now the situation in which, in addition to the unlabeled samples from both domains, we also have small datasets of labeled pairs $\{(x_i^1, y_i^1)\}$ and $\{(x_i^2, y_i^2)\}$. Our goal is to utilize unsupervised learning by exploiting the shared features joint autoencoders extract. We describe two strategies to achieve that goal using joint autoencoders.

The first approach, particularly relevant to transfer learning, comprises several steps. First, we train joint autoencoders on both tasks simultaneously, using all available unlabeled data. Then, for the source task (the one with more labeled examples), we fine-tune the branches up to the shared layer using the set of labeled pairs, and freeze the learned shared layers. Finally, for the target task, use the available labeled data to train only its private branches while fixing the, shared layers fine-tuned on the source data.

The second, *end-to-end* approach, combines supervised and unsupervised training. Here we extend the JAE architecture by adding new layers, with supervised loss functions for each task; see Fig 2. We train the new network using all losses from all tasks simultaneously - reconstruction losses using unlabeled data, and supervised losses using labeled data. When the size of the labeled sets is highly non-uniform, the network is naturally suitable for transfer learning. When the labeled sample sizes are roughly of the same order of magnitude, the setup is suitable for semi-supervised learning.

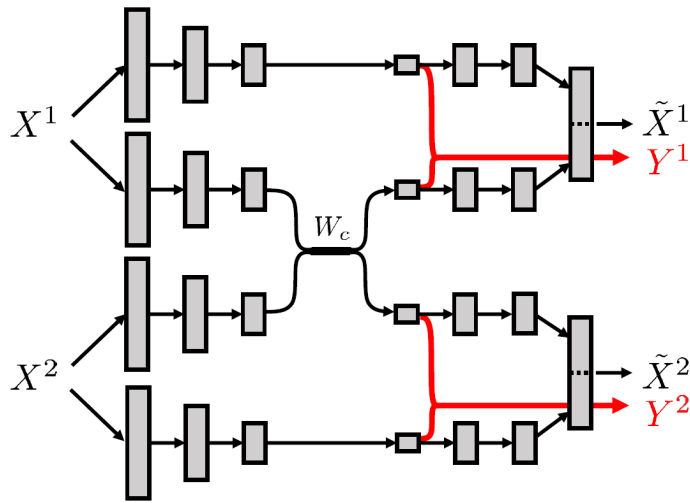

Figure 2: A schematic depiction of a joint autoencoder architecture extended for supervised learning.

### 3.3 On the depth of sharing

It is common knowledge that similar *low-level features* are often optimal for similar tasks. For example, in many vision applications, convolutional neural nets exhibit the same Gabor-type filters in their first layer, regardless of the specific objects they are trained to classify. This observation makes low-level features immediate candidates for sharing in multi-task learning settings. However, unsurprisingly, sharing low-level features is not as beneficial when working with domains of different nature (e.g., handwritten digits vs. street signs).

In this paper, we propose to share weights in rather deeper layers of a neural net, while leaving the shallow layers un-linked. The key idea is that by forcing two nets to use the same deep weights, their preceding shallow layers must learn to transform the data from the two domains into a common form. We support this intuition through several experiments. As our preliminary results in Section 4.2.1 show, for similar domains, sharing deep layers provides the same performance boost as sharing shallow layers. Thus, we do not pay a price for relying only on "deep similarities". But for domains of a different nature, sharing deep layers has a clear advantage.

## 4 Experiments

All experiments were implemented in Keras [8] over Tensorflow [1]. The code will be made available soon. All the images were scaled to $[0, 1]$. Unless specified otherwise, the ADAM optimizer was used.

### 4.1 Unsupervised learning

We present experimental results demonstrating the improvement in unsupervised learning of multiple tasks on the MNIST [24] and CIFAR-10 [23] datasets. For the MNIST experiment, we have separated the training images into two subsets: $X^1$, containing the digits $\{0, 1, 2, 3, 4\}$ and $X^2$, containing the digits $\{5, 6, 7, 8, 9\}$. We compared the $L_2$ reconstruction error achieved by the JAE portrayed in Figure 1(b) to a baseline of a pair of standard autoencoders trained on each dataset with architecture identical to a single branch of the JAE (Fig 1(a)). The joint autoencoder (MSE =5.4) out-performed the baseline (MSE = 5.6) by 4%.

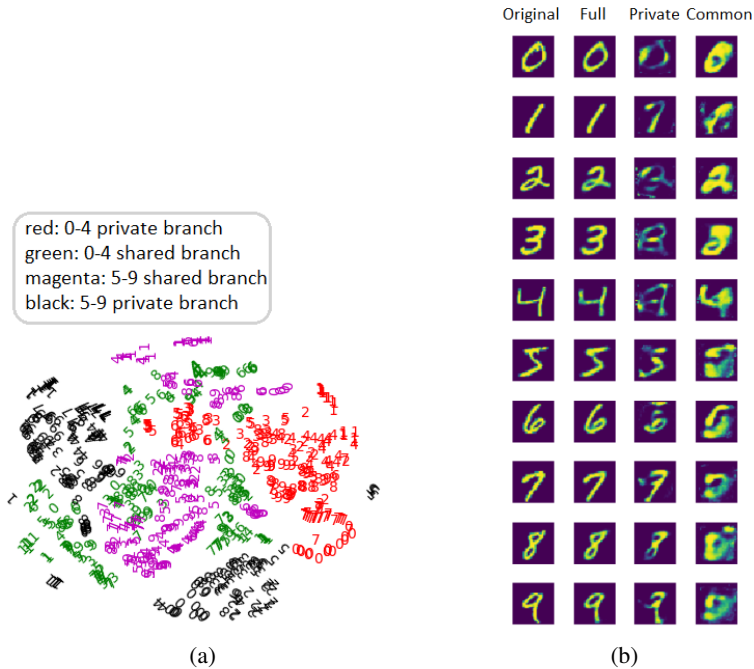

(a)                              (b)

Figure 3: (a) T-sne visualizations of the responses of each bottleneck to images from $\{0-4\}$ and $\{5-9\}$ MNIST digits: red and black for the private branches of the datasets, green and magenta for the shared branches. (b) From left to right: original digits, reconstruction by the JAE, reconstruction by the private branch, reconstruction by the shared branch.

To further understand the features learned by the shared and private bottlenecks, we visualize the activations of the bottlenecks on $1000$ samples from each dataset, using $2D$ t-sne embeddings [36]. Figure 3(a) demonstrates that the common branches containing the shared layer (green and magenta) are much more mixed between themselves than the private branches (red and black), indicating that they indeed extract shared features. Figure 3(b) displays examples of digits reconstruction. The columns show (from left to right) the original digit, the image reconstructed by the full JAE, the output of the private branches and the shared branches. We see that the common branches capture the general shape of the digit, while the private branches capture the fine details which are specific to each subset.

We verify quantitatively the claim about the differences in separation between the private and shared branches. The Fisher criterion [7] for the separation between the t-sne embeddings of the private branches is $7.22 \cdot 10^{-4}$, whereas its counterpart for the shared branches is $2.77 \cdot 10^{-4}$, 2.6 times less. Moreover, the shared branch embedding variance for both datasets is approximately identical, whereas the private branches map the dataset they were trained on to locations with variance greater by $1.35$ than the dataset they had no access to. This illustrates the extent to which the shared branches learn to separate both datasets better than the private ones.

For CIFAR-10, containing images from 10 classes, we trained the baseline autoencoder on single-class subsets of the database (e.g., all airplane images) and trained the joint autoencoder on pairs of such subsets. Table 1 shows a few typical results, demonstrating a consistent advantage for JAEs over standard autoencoders. Besides the lower reconstruction error evident in each experiment, we can see that classes of images that are visually similar, enjoy a greater boost in performance. For instance, the pair deer-horses enjoyed a performance boost of $37\%$, which is greater than the typical boost of 33-35%.

Table 1: JAE reconstruction performance

|              | A-D  | A-H  | A-S  | D-H  | D-S  | H-S  |
|--------------|------|------|------|------|------|------|
| AE error     | 20.8 | 18.5 | 16.2 | 20.6 | 18.2 | 16.0 |
| JAE error    | 13.9 | 12.2 | 10.8 | 13.2 | 11.4 | 10.6 |
| Improvement  | 33%  | 34%  | 33%  | 37%  | 35%  | 34%  |

Performance of JAEs vs standard autoencoders in terms of reconstruction MSE on pairs of objects in CIFAR-10: airplanes (A), deer (D), horses (H), ships(S). Training was performed using SGD, with learning rate $0.001$, batch size $128$, weight decay $5 \cdot 10^{-4}$. For each pair of objects, we give the standard AE error, JAE error and the relative improvement percentage.

### 4.2 Transfer learning

First, we compare the performance of the two JAE-based nearly-unsupervised transfer learning methods detailed in Section 3.2. For both methods, $X^1$ contains digits from $\{0, 1, 2, 3, 4\}$ and $X^2$ contains the digits $\{5, 6, 7, 8, 9\}$. The source and target datasets comprise $2000$ and $500$ samples, respectively. Both methods are compared against a simple neural network with two fully-connected layers of dimensions 60 and 30, trained on the target samples alone. All results are measured on the MNIST test set. Table 2 presents the results of this comparison, demonstrating the improved performance with respect to the baseline.

Table 2: JAE Transfer learning methods comparison

| Method                     | $X^1 \to X^2$ | $X^2 \to X^1$ |
|----------------------------|---------------|---------------|
| Baseline                   | 89.4          | 95.8          |
| Common branch transfer     | 92.3          | 96.1          |
| End-to-end with supervised | 96.6          | 98.3          |

Comparison of a simple baseline trained on target samples, common branch transfer and end-to-end semi-supervised methods on MNIST digits {0-4} and {5-9}.

#### 4.2.1 Shared layer depth

We investigate the influence of the depth of the shared layers on the transfer performance. Table 3 presents the results of this study. It can be seen that for highly similar pairs of tasks such as the two halves of the MNIST dataset, the depth is of little significance, while for dissimilar pairs such as MNIST-USPS, "deeper is better" - the performance improves with the shared layer depth. Moreover, when the input dimensions differ, early sharing is impossible - the data must first be transformed to have the same dimensions.

#### 4.2.2 MNIST, USPS and SVHN digits datasets

We have seen that the end-to-end JAE-with-transfer algorithm outperforms the alternative approach. We now compare it to other domain adaptation methods from [34] that use little to no target samples for supervised learning, applied to the MNIST, USPS [21] and SVHN [32] digits datasets. The transfer tasks we consider are MNIST→USPS, USPS→MNIST and SVHN →MNIST. Following [34] and [28], we use 2000 samples for MNIST and 1800 samples from USPS. For the SVHN→MNIST adaptation, we use the complete training sets. In all three tasks, both the source and the target samples are used for the unsupervised JAE training. In addition, the source samples are used for the source supervised element of the network. For the MNIST vs. USPS tasks, 200 target

Table 3: Shared Layer Depth and Transferability

|  | 1 | 2 | 3 | 4 | 5 |
|---|---|---|---|---|---|
| MNIST {0-4} → {5-9} | 0.965 | 0.954 | 0.958 | 0.961 | 0.960 |
| MNIST {5-9} → {0-4} | 0.983 | 0.976 | 0.978 | 0.982 | 0.983 |
| MNIST → USPS |  |  | 0.848 |  | 0.876 |
| USPS → MNIST |  |  | 0.832 |  | 0.869 |

Influence of the shared layer depth on the transfer learning performance. For the MNIST-USPS pair, only partial data are available for dimensional reasons.

samples (20 per digit) are also used for supervised fine-tuning on the target. When performing the SVHN→MNIST transfer learning, we consider two protocols. In one, 20 target samples per digit are used, while in the other, only 5 target samples per digit are available for the supervised training.

Table 4 provides the results of our experiments. On all tasks, we achieve results comparable or superior to existing methods, despite JAE being simpler than most competing approaches. In particular, we do not train a GAN. The SVHN→MNIST task is considered to be the hardest (for instance, GAN-based approaches fail to address it) yet the abundance of unsupervised training data allows us to achieve the best results, relative to the existing methods.

Table 4: Transfer learning results on MNIST, USPS and SVHN

| Method | MNIST → USPS | USPS → MNIST | SVHN → MNIST |
|---|---|---|---|
| Gradient reversal | $0.771 \pm 0.018$ | $0.730 \pm 0.020$ | 0.739 |
| Domain confusion | $0.791 \pm 0.005$ | $0.695 \pm 0.033$ | $0.681 \pm 0.003$ |
| CoGAN | $0.912 \pm 0.008$ | $0.891 \pm 0.008$ | did not converge |
| ADDA | $0.894 \pm 0.002$ | $0.901 \pm 0.008$ | $0.760 \pm 0.0018$ |
| **JAE (Ours)** | $\mathbf{0.876 \pm 0.032}$ | $\mathbf{0.869 \pm 0.052}$ | $\mathbf{0.805 \pm 0.003}$ (5 samples) $\mathbf{0.918 \pm 0.011}$ (20 samples) |

## 5  Conclusion

We presented a general scheme for incorporating prior knowledge within deep feedforward neural networks for domain adaptation, multi-task and transfer learning problems. The approach is general and flexible, operates in an end-to-end setting, and enables the system to self-organize to solve tasks based on prior or concomitant exposure to similar tasks, requiring nothing more than standard gradient based optimization for learning. The basic idea of the approach is the sharing of representations for aspects which are common to all domains/tasks while maintaining private branches for task-specific features. The method is applicable to data from multiple sources and types, and has the advantage of being able to share weights at arbitrary levels of the network, enabling abstract levels of sharing.

We demonstrated the efficacy of our approach on several domain adaptation and transfer learning problems, and provided intuition about the meaning of the representations in various branches. In a broader context, it is well known that the imposition of structural constraints on neural networks, usually based on prior domain knowledge, can significantly enhance their performance. The prime example of this is, of course, the convolutional neural network. Our work can be viewed within that general philosophy, showing that improved functionality can be attained by the modular prior structures imposed on the system, while maintaining simple learning rules.

# Appendix

**A. Auto-Encoder Architectures Employed in the Paper**

In the experimental part of the paper, a variety of autoencoders is used. Whenever possible, we strive to utilize existing architectures, common in the community. Below is a brief description, for the sake of completeness. For each dataset (CIFAR-10, USPS, SVHN), we describe the original autoencoder, as well the JAE branches. For the first two datasets, we describe only a single JAE branch, because they all have the same architecture.

Notation (for non-obvious layers): $Conv(k, k, n)$ stands for a convolution layer with $n$ filters of size $k \times k$, $DeConv(k, k, n)$ stands for a convolution layer with $n$ filters of size $k \times k$ with $2 \times 2$ upsampling. $MP$ stands for max-pooling of size $2 \times 2$ and $Merge$ stands for merging with a parallel layer from another branch. Finally, $FC - k$ represents a fully-connected layer of size $k$.

CIFAR-10 autoencoder: $Conv(5, 5, 32)$, $ReLU$, $MP$, $Conv(5, 5, 96)$, $ReLU$, $MP$, $DeConv(2, 2, 32)$, $DeConv(2, 2, 3)$, $Sigmoid$

CIFAR-10 JAE branch: $Conv(5, 5, 32)$, $ReLU$, $MP$, $Conv(5, 5, 48)$ (shared), $ReLU$, $MP$, $DeConv(2, 2, 16)$, $Merge$, $DeConv(2, 2, 3)$, $Sigmoid$

USPS autoencoder: $Conv(5, 5, 12)$, $ReLU$, $MP$, $Conv(3, 3, 30)$, $ReLU$, $Conv(3, 3, 100)$, $Flatten$, $FC - 250$, $FC - 60$, $FC - 100$, $FC - 300$, $FC - 256$, $Sigmoid$

USPS JAE branch: $Conv(5, 5, 12)$, $ReLU$, $MP$, $Conv(3, 3, 30)$, $ReLU$, $Conv(3, 3, 100)$, $Flatten$, $FC - 250$, $FC - 30$ (shared), $FC - 100$, $FC - 300$, $FC - 128$, $Merge$, $FC - 256$, $Sigmoid$

SVHN autoencoder: $Conv(5, 5, 20)$, $ReLU$, $MP$, $Conv(5, 5, 50)$, $ReLU$, $Conv(5, 5, 500)$, $Flatten$, $FC - 250$, $FC - 500$, $FC - 3072$, $Sigmoid$

SVHN private branch: $Conv(5, 5, 20)$, $ReLU$, $MP$, $Conv(5, 5, 50)$, $ReLU$, $Conv(5, 5, 500)$, $Flatten$, $FC - 250$, $FC - 220$, $FC - 250$, $Merge$, $FC - 3072$, $Sigmoid$

SVHN shared branch: $Conv(5, 5, 20)$, $ReLU$, $MP$, $Conv(5, 5, 50)$, $ReLU$, $Conv(5, 5, 500)$, $Flatten$, $FC - 250$, $FC - 30$ (shared), $FC - 250$, $Merge$, $FC - 3072$, $Sigmoid$